\documentclass[times,twocolumn,final,authoryear]{elsarticle}

\usepackage{prletters}
\usepackage{framed,multirow}

\usepackage{amssymb}
\usepackage{latexsym}
\usepackage{amsmath}

\usepackage{url}
\usepackage{xcolor}
\definecolor{newcolor}{rgb}{.8,.349,.1}

\journal{Pattern Recognition Letters}

\begin{document}

\thispagestyle{empty}

\clearpage

\ifpreprint
  \setcounter{page}{1}
\else
  \setcounter{page}{1}
\fi

\begin{frontmatter}

\title{Guiding Intelligent Surveillance System by learning-by-synthesis gaze estimation}

\author[1]{Tongtong \snm{Zhao}}

\author[1]{Yuxiao \snm{Yan}}
\author[1]{Jinjia \snm{Peng}}
\author[1]{Zetian \snm{Mi}}
\author[1]{Xianping \snm{Fu}\corref{cor1}}

\cortext[cor1]{Corresponding author:
	Tel.: +0-000-000-0000;
	fax: +0-000-000-0000;}
\ead{fxp@dlmu.edu.cn}

\address[1]{Information Science and Technology College, Dalian Maritime University, Dalian, China.}

\received{}
\finalform{}
\accepted{}
\availableonline{}
\communicated{}

\begin{abstract}
We describe a novel learning-by-synthesis method for estimating gaze direction of an automated intelligent surveillance system. Recently, progress in learning-by-synthesis has proposed training models on synthetic images, which can effectively reduce the cost of manpower and material resources. However, learning from synthetic images still fails to achieve the desired performance compared to naturalistic images due to the different distribution of synthetic images. In an attempt to address this issue, previous method is to improve the realism of synthetic images by learning a model. However, the disadvantage of the method is that the distortion has not been improved and the authenticity level is unstable. To solve this problem, we put forward a new structure to improve synthetic images, via the reference to the idea of style transformation, through which we can efficiently reduce the distortion of pictures and minimize the need of real data annotation. We estimate that this enables generation of highly realistic images, which we demonstrate both qualitatively and with a user study. We quantitatively evaluate the generated images by training models for gaze estimation. We show a significant improvement over using synthetic images, and achieve state-of-the-art results on various datasets including MPIIGaze dataset.
\end{abstract}

\begin{keyword}
Generative Adversarial Networks(GANs) \sep Style Transfer  \sep Learning-by-synthesis
\end{keyword}

\end{frontmatter}


\section{Introduction}
In ordinary day-to-day behaviour humans identify the intentions of others by drawing on knowledge of the world that they have accumulated throughout their lifetime. On the contrary, for intelligent surveillance system (\cite{hu-car}), world knowledge is very limited, thus making it very difficult to make such inferences. Eyes and their movements can represent feelings and desire, reveal human attention and play an important role in social communication. Therefore, gaze estimation method becomes an effective means to guide the intelligent surveillance system to recognize \cite{wang2016semantic,feng2018learning} the personal intention. We can capture people's attention priority via gaze estimation technology. Furthermore,it makes the judgement of people's criminal intent effective.

There is no denying that, in recent years, gaze estimation has been able to meet the needs of actual landing scenarios such as intelligent surveillance system under the training of a large amount of data. However, due to the high cost of time and bankroll, solutions are required to tackle these problems. When it comes to this matter, human give priority to the synthetic image because the annotations are automatically available. However, learning the misleading synthetic images cause owing to the gap between synthetic and real image distributions-synthetic data is not the copy of the realism, the details represented confuse the network and render it fail to complete the mission.\par
As such, one solution is to improve the simulator. But increasing the authenticity is computationally expensive, designing a renderer is a heavy workload, and the top renderer may still be difficult to model all the features of the real image. This may make the model over fitting in the "unreal" details of the synthetic image. The other solution is to improve the distribution of synthetic images and make them closer to the real pictures. The current method of state-of-the-art is \cite{DBLP:journals/corr/ShrivastavaPTSW16}. We adopt a neural network model similar to Generative Adversarial Networks (GAN). The main use of GAN was to train computers to generate some emanational pictures. To be graphic, it uses a synthetic-image-producing network to be against another dataset that produces real pictures, and then distinguish it with a separate distinction network. On the base of GAN, they make some big difference on models. For example, they input synthetic images instead of random vectors and propose a learning model called ¡®Simulated + Unsupervised¡¯ ultimately.\par
The contribution of this paper to computer vision, in addition to a new learning model, also includes using the model successfully train an optimized network (Refiner) on the premise of no artificial annotation and rendering computers generate more ¡®real¡¯ synthetic images. However, the disadvantage of the method is that the distortion is not improved and the authenticity level is not stable. So, to solve this problem, we put forward a new structure, which can improve synthetic images, via the reference to the idea of style transformation to efficiently reduce the distortion of pictures and minimize the need of real data annotation. The same as general GAN structure, our framework also includes the generation network G and the distinction network D. We improve the structure of the image generation part and change the input from the random vector to the content of real image distribution and the simulation picture together. It will make the generation more stable, avoiding the randomness of distribution. It will also achieve a stable distribution in a short time. We modify the way of loss evaluating of the distinction network and add regular items to ensure the authenticity of the pictures.\par
In summary, our contributions are five-fold:\par
1.	We propose a new structure, which can improve synthetic images, via the reference to the idea of style transformation to efficiently reduce the distortion of pictures and minimize the need of real data annotation.\par
2.	We improve the structure of the image generation part and  modify the way of loss evaluating of the distinction network and add regular items to ensure the authenticity of the pictures. It will make the generation more stable, avoiding the randomness of distribution. \par
3.	We performance experiments to verify proposed structure can generate highly realistic images steadily by qualitative and user research. Meanwhile, the training model of gaze estimation is used to evaluate produced images quantitatively. Compared with the synthetic images used, we implemented the best results on multiple datasets.\par

\section{Related Works}

The most prominent contemporary approach to refine synthetic images (change the distribution of synthetic images) is based on generative adversarial networks (GANs).The GANs framework learns a generator and a discriminator with competing losses. The goal of generator is to map the a random vector to a realistic image,whereas the goal of the discriminator is to distinguish the generated and the real images. In the original work of \cite{DBLP:journals/corr/GoodfellowPMXWOCB14}, GANs (\cite{DBLP:journals/corr/GoodfellowPMXWOCB14}) were used to generate visually realistic images. Since then ,many improvements have been proposed to realism synthetic images. \cite{SGAN} used a Structures GANs to learn surface normals and then combine it with a Style GANs to generate natural indoor scenes. \cite{Perceptual} introduced a family of composite loss functions for image synthesis, which combined regression over the activations of a fixed perceiver network with a GANs (\cite{DBLP:journals/corr/GoodfellowPMXWOCB14}) loss. \cite{DeepFont} trained a Stacked Convolutional Auto-Encoder on synthetic and real data to learn the low-level representations of their font detector ConvNet.

\cite{Wang2015Robust} presented a novel approach towards subspace clustering over multi-view data(\cite{Wang2015Effective,wang2018beyond,wang2016multi}), and further not only proposed an iterative structured low-rank optimization method to multi-view spectral clustering (\cite{Wang2016Iterative} \cite{Wang2018Multiview}), but also a collaborative deep network for robust landmark retrieval(\cite{Wang2017Effective}). \cite{Wu2017Deep} proposed a principled deep feature embedding approach for person reidentification and  presented a novel deep attention-based spatially recursing model for fine-grained visual recognition(\cite{Wu2018Deep,wu2018and,wu2018whereand}).

The most relevant to our work is \cite{DBLP:journals/corr/ShrivastavaPTSW16} which propose Simulated+Unsupervised (S+U) learning, where the task is to learn a model to improve the realism of a simulator$¡¯$s output using unlabeled real data, while preserving the annotation information from the simulator. They develop a method for S+U learning that uses an adversarial network similar to Generative Adversarial Networks (GANs), but with synthetic images as inputs instead of random vectors. Similar with \cite{DBLP:journals/corr/ShrivastavaPTSW16}, we also uses an adversarial network similar to Generative Adversarial Networks (GANs) to refine the synthetic images, but we improve the structure of the image generation part and change the input from the random vector to the content of real image distribution and the simulation picture together. It will make the generation more stable, avoiding the randomness of distribution. It will also achieve a stable distribution in a short time. Besides that, we modify the way of loss evaluating of the distinction network and add regular items to ensure the authenticity of the pictures.

Style transfer algorithms is another way to change the distribution of images. Global style transfer algorithms process an image by applying a spatially-invariant transfer function. These methods are effective and can handle simple styles like global color shifts (e.g., sepia) and tone curves (e.g., high or low contrast). For instance, Reinhard et al. match the means and standard deviations between the input and reference style image after converting them into a decorrelated color space. Local style transfer algorithms based on spatial color mappings are more expressive and can handle a broad class of applications. For instance, \cite{Tatarchenko} train a ConvNet to generate images of 3D models, given a model ID and viewpoint. The network thus acts directly as a rendering engine for the 3D
model. ¡±pix2pix¡± of \cite{Conditional}, which uses a conditional GAN to learn a mapping from input to output images. Similar ideas have been applied to various tasks such as generating photographs from sketches or from attribute and semantic layouts (\cite{Karacan}). Unlike the earlier work, our approach improve synthetic images, via the reference to the idea of style transformation to efficiently reduce the distortion of pictures and minimize the need of real data annotation.

\section{Proposed Method}\label{sec:Proposedmethod}

\begin{figure}[!t]
\centering
\includegraphics[scale=.5]{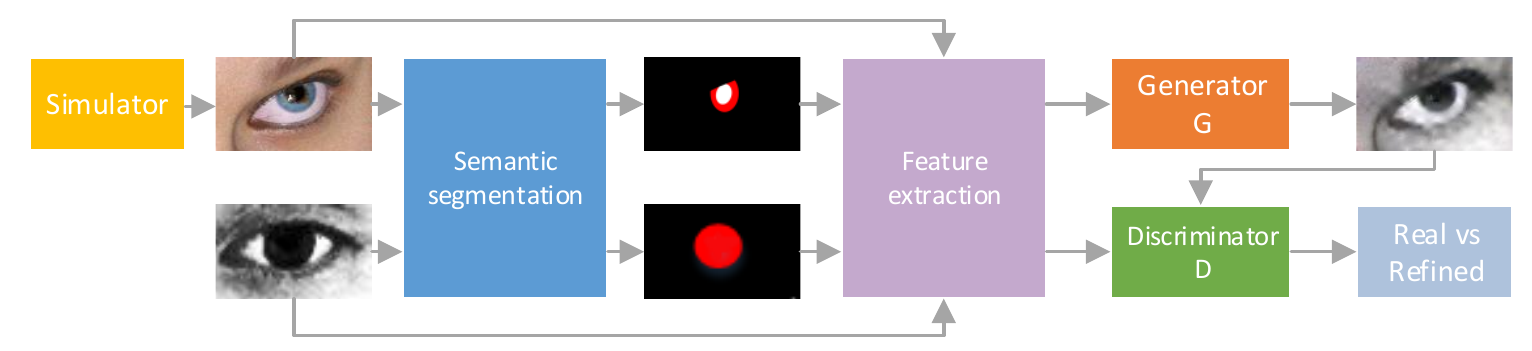}
\caption{The overview of proposed methods.}
\end{figure}

Our proposed network (As Fig.1) takes two images with their mask: the reference style image which is a set of naturalistic eye image from video of driving environment or naturalistic eye image dataset. A stylized and retouched image referred as the input image from synthetic image dataset. We use this to train the gaze estimation, as we seek to transfer the style of the reference to the input while keeping the content and spatial information due to its importance in appearance-based gaze estimation. The proposed network can be divided into four parts: coarse segmentation network, feature extraction network, Generator and Discriminator.

We train the semantic segmentation network which builds upon an efficient redesign of convolutional blocks with residual connections to segment, according to the line of gaze estimation for the naturalistic image. One of the great benefits of synthetic data is that its semantic information is clearer. Thus the challenge is mainly on segment naturalistic image. Residual connections can avoid the degradation problem with a large amount of stacked layers. Our architecture is fully depicted in fig.2. $Number$ $of$ $feature$ $maps$ $at$ $layers$ $@$ $output$ $resolution$ is shown under each block.

\begin{figure}[!htbp]
\centering
 \includegraphics[width = 0.5\textwidth,angle=0]{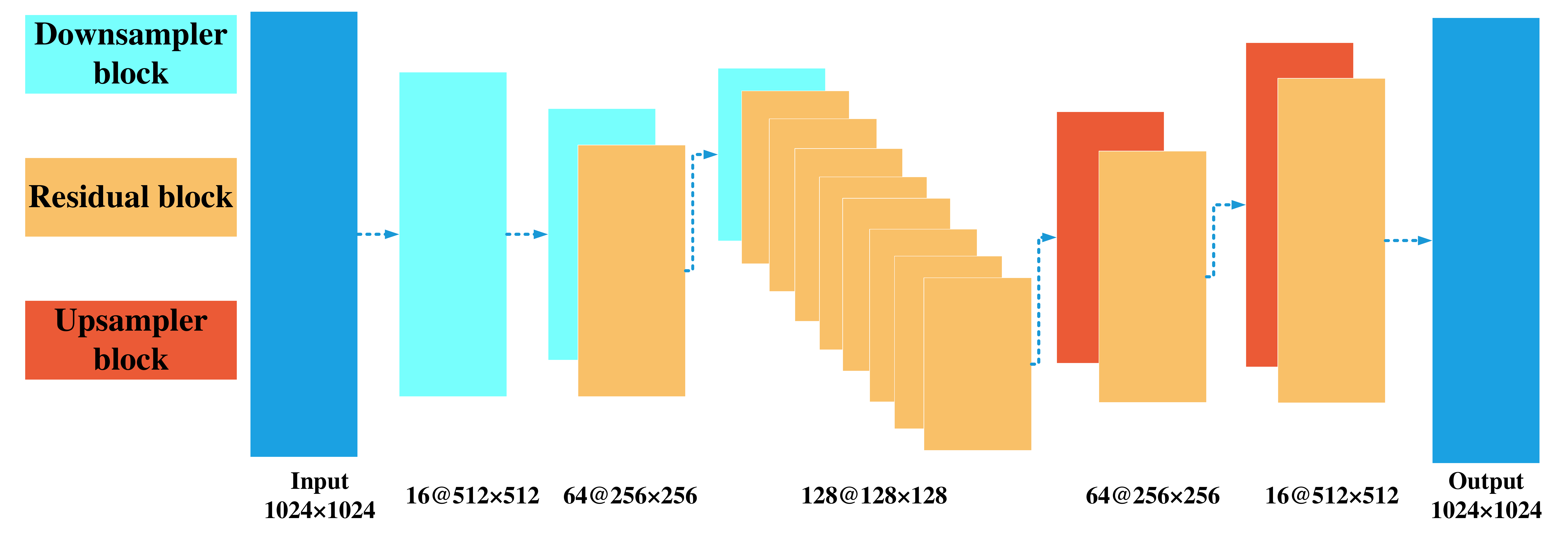}
    \caption{The overview of semantic segmentation network.$Number$ $of$ $feature$ $maps$ $at$ $layers$ $@$ $output$ $resolution$ is shown under each block. The network has three kinds of block. The structure of these block is shown in fig.2. We follow an encoder-decoder architecture to avoid the need of using skip layers to refine the output. Furthermore, in consideration of simplifying the task, we only mark two kinds of information on the naturalistic image: the pupil and the iris.}
    \label{figotherdatabase}
\end{figure}

 As we know, Residual block consist of many stacked ¡°Residual Units¡± and each unit can be expressed in a general form as $y_{l}=h(x_{l})+F(x_{l},W_{l}$,$x_{l+1}=f(y_{l})$ where $x_{l}$ and $x_{l+1}$ are input and output of the $l-th$ unit, and F is a residual function. In $y_{l}=h(x_{l})+F(x_{l},W_{l}$,$x_{l+1}=f(y_{l})$, $h(x_{l})=x_{l}$ is an identity mapping and f is a ReLU function. We try to change the residual network structure makes the association between features stronger. Furthermore, in consideration of simplifying the task, we only mark two kinds of information on the naturalistic image: the pupil and the iris. However, many naturalistic images are influenced by light and other factors, and sometimes the pupil and the iris cannot be completely separated, to avoid "orphan semantic labels" that are only present in the input image, which the "orphan labels" usually are pupil region because of the outdoor illumination effect, we constrain the pupil semantic region to be set as the center of iris region. We have also observed that the segmentation does not need to be pixel accurate since eventually, the output is constrained by feature extraction network.

\subsection{Feature Extraction network}
The architecture of the feature extraction network is shown as Fig.{3}, the network has an encoder-decoder structure with skip connections. To ensure the features are consistent within each instance, we add an instance wise average pooling layer to the putput of the encoder to compute the average feature for the instance. The decoder uses the representation to synthesize progressively finer feature maps.

$\mathbf{Encoder.}$ Our encoder is based on VGG-19. The network consists of five models and each module contains a number of convolutional layers with layer normalization, ReLU and average pooling. The first module has two convolutional layers, while each of the other modules have three.

$\mathbf{Decoder.}$ Our decoder is based on the cascaded refinement network (CRN). The network is a cascade of refinement modules. Each refinement module contains two convolutional layers with layer normalization and Leaky ReLU.

\begin{figure}[!htbp]
\centering
 \begin{minipage}[]{0.5\textwidth}
    \centering
     \includegraphics[width = 1\textwidth,angle=0]{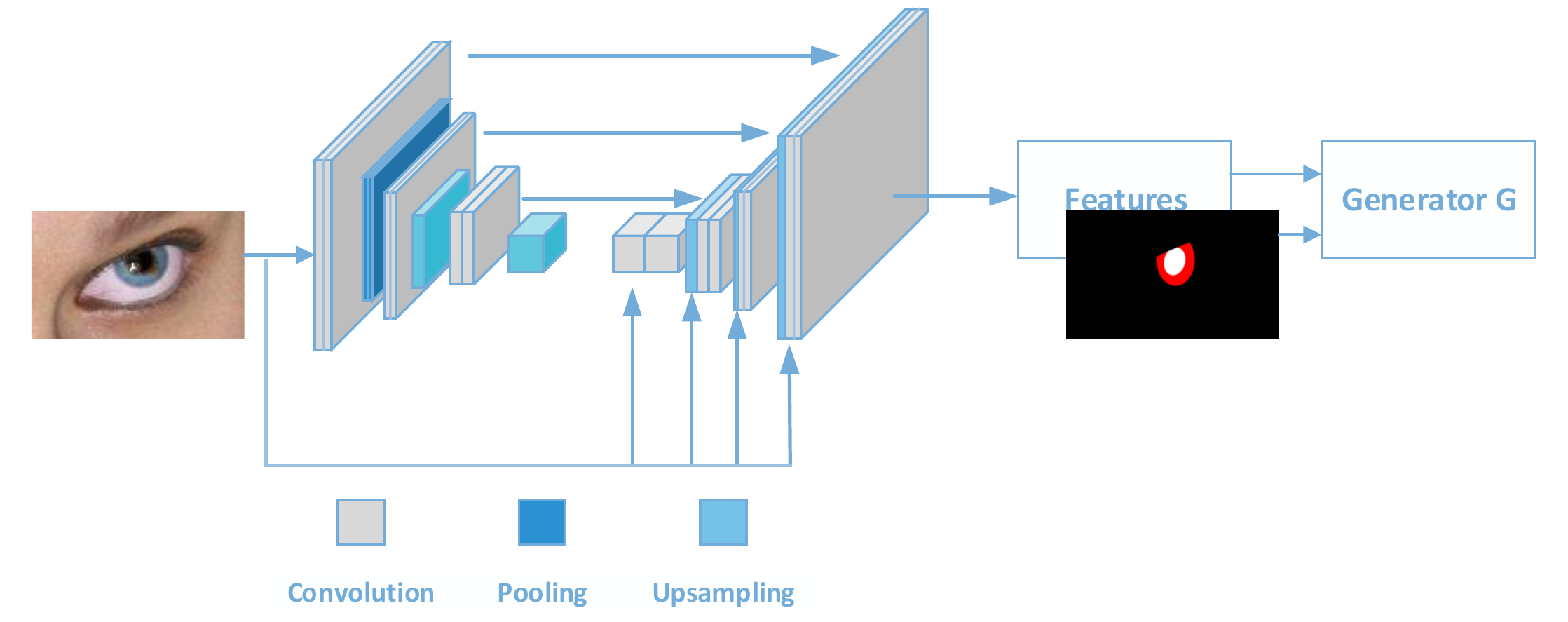}
 \end{minipage}
    \caption{Using instance-wise features in addition to labels for generating images. }
    \label{figfeature}
\end{figure}

\subsection{Generator G}
We decompose the generator into two-subnetworks:G1 and G2. We term G1 as the global generator network and G2 as the local enhancer network. The generator is then given by the tuple $G={G1,G2}$ as visualized in Fig.5. The global generator The global generator network
operates at a resolution of $297 \times 297$, and the local enhancer network outputs an image with a semantic layouts that is the output of the previous semantic segmentation network (\cite{deng2018learning}).

Our global generator is built on the architecture proposed by Johnson et al. [22], which has been proven successful for neural style transfer on images. It consists of 3 components: a convolutional front-end G1(F), a set of residual blocks G1(R) and a transposed convolutional back-end G1(B).

The local enhancer network also consists of 3 components: a convolutional front-end G2(F) , a set of residual blocks G2(R), and a transposed convolutional back-end G2(B). Different from the global generator network, a semantic label map  is passed through the 3 components sequentially to output an image with instance segmentation information and the input to the residual block G2(R) is the element-wise sum of two feature maps: the output feature map of G2(F) , and the last feature map of the back-end of the global generator
network G1(B). This helps integrating the global information from G1 to G2.

During training, we first train the global generator and then train the local enhancer in the order of their scale. We then jointly fine-tune all the networks together. We use this generator design to effectively aggregate global and local information for the image synthesis task.

\begin{figure}[!t]
\centering
\includegraphics[scale=.5]{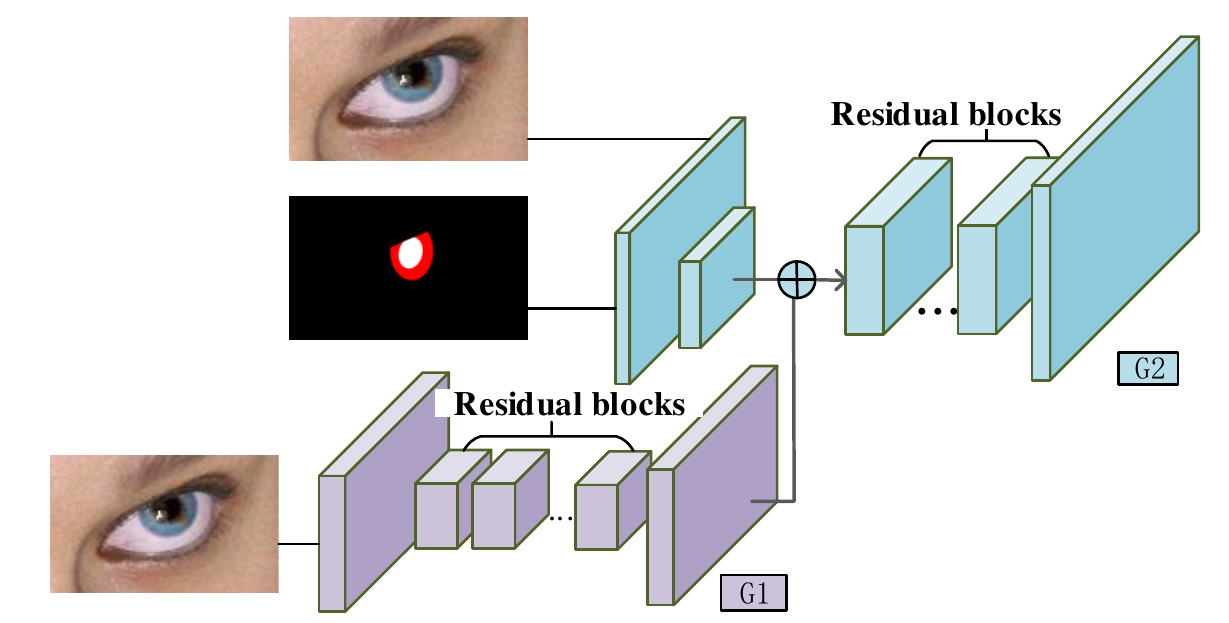}
\caption{Network architecture of our generator.}
\end{figure}

\subsection{Discriminator D}

Realistic image synthesis poses a great challenge to the GAN discriminator design. To differentiate distribution real and synthesized
images, the discriminator needs to have a large receptive field with instance segmentation information on global and local images. This would require either a deeper network or larger convolutional kernels. As both choices lead to an increased network capacity, overfitting would become more of a concern. Also, both choices require a larger memory footprint for training, which is already a scarce resource for realistic image generation. Inspired by Style Transfer, we propose Discriminator D with novel loss function which is a pretrained VGG-19 (\ cite{Simonyan2014})  network and made some key modifications to the standard perception losses to keep the distribution of the naturalistic images and content of the synthetic images to the fullest extent. As Fig.5 shows that instead of taking only RGB color channels into consideration, our network utilizes the representations of both color and semantic features for style transfer. With the semantic features, we can address the spatial arrangement information and avoid the spatial configuration of the image being disrupted because of the style transformation.

Feature Gram matrices are effective at representing texture, because they capture global statistics across the image due to spatial averaging. Since textures are static, averaging over positions is required and makes Gram matrices fully blind to the global arrangement of objects inside the reference real image. So if we want to keep the global arrangement of objects, make the gram matrices more controllable to compute over the exact region of entire image, we need to add some texture information to the image. \cite{Deep2017} present a method which add the masks to the input image as additional channels and augment the neural style algorithm by concatenating the segmentation channels, inspired by it, mask is added as the texture information we need to compute over the exact region of entire image, thus the style loss can be denoted as:
\begin{equation}
\ell_{style}^l=\lambda_{g}\ell_{gs}^l+\lambda_{l}\ell_{ls}^l
\end{equation}
 \begin{equation}
\ell_{gs}^l=\sum_{c=1}^{C}\frac{1}{4N^2_{l,c}M^2_{l,c}}\sum_{ij}\left(G_{l}[O]-G_{l}[S]\right)^2_{ij}
\end{equation}
\begin{equation}
\ell_{ls}^l=\sum_{c=1}^{C}\frac{1}{4N^2_{l,c}M^2_{l,c}}\sum_{ij}\left(G_{l,c}[O]-G_{l,c}[S]\right)^2_{ij}
\end{equation}

where  C is the number of channels in the semantic segmentation mask and $l$ indicates the $l$-th convolutional layer of the deep convolutional neural network. Each layer with $N_{l}$ distinct filters has $N_{l}$ feature maps each of size $M_{l}$, where $M_{l}$ is the height times the width of the feature map. So the responses in each layer $l$ can be stored in a matrix $F[\cdot] \in R^{N_{l}\times M_{l}}$ where $F[\cdot]_{ij}$ is the activation of the $i^{th}$ filter at position $j$ in each layer $l$.
\begin{equation}
F_{l,c}[O]=F_l[O]S_{l,c}[I]
\end{equation}
\begin{equation}
F_{l,c}[S]=F_l[S]S_{l,c}[S]
\end{equation}
\begin{equation}
G_{l,c}[\cdot]=F_{l,c}[\cdot]F_{l,c}[\cdot]^{T}
\end{equation}
$S_{l,c}[\cdot]$ is the segmentation mask in each layer $l$ with the channel c. $\lambda_{g}$ is the weight to configure layer preferences of global losses $\ell_{gs}$ which calculated between raw input image and features which was extracted by feature extraction network.$\lambda_{l}$ is the weight to configure layer preferences of local losses $\ell_{ls}$ which calculated between input segmentation image and features which was extracted by feature extraction network with the input of segmentation image.

We now describe how we regularize this optimization scheme to preserve the structure of the input image and produce realistic but no distorted outputs. Our strategy is to express this constraint not on the output image directly but on the transformation that is applied to the input image. We name $Vc[O]$ the vectorized version ($N \times 1$) of the output image O in channel c and define the following regularization term that penalizes outputs that are not well explained by a locally affine transform:
\begin{equation}
\ell_{m} = \sum_{c=1}^{3}Vc[O]^{T}Vc[O]
\end{equation}
We formulate the realistic but no distorted style transfer objective by combining all 3 components together:
\begin{equation}
L_{total}=\eta\sum_{l=1}^L\beta_l\ell_{style}^l + \vartheta\ell_{m}
\end{equation}
where $\eta=10^{2}$,$\vartheta=10^{4}$

Our full objective combines both GAN loss $\ell_{GAN}$ and style tranfer loss $L_{total}$ as:
\begin{equation}
min_{G}(\sum \ell_{GAN}(G,\ell^{l}) + \lambda\sum L_{total})
\end{equation}
where $\lambda$ controls the importance of the two terms.

\section{Experimental Results}\label{sec:experimentalresults}

\subsection{Implementation Details}
This section describes the implementation details of our approach. We employed the pre-trained VGG-19 as the
feature extractor. We chose conv$4_2$($\alpha_l=1$ for this layer,$\alpha_l=0$ for other layers) as the local content representation, and conv$1_1$, conv$2_1$, conv$3_1$, conv$4_1$, and conv$5_1$($\beta_{l}=\frac{1}{5}$ for these five layers,$\beta_{l}=0$ for all other layers) as the local style representation. conv$3_2$ ($\alpha_l=1$ for this layer,$\alpha_l=0$ for other layers)  as the global content representation, and conv$1_2$, conv$2_2$, conv$3_3$, conv$4_3$, and conv$5_3$ ($\beta_{l}=\frac{1}{5}$ for these five layers,$\beta_{l}=0$ for all other layers) as the global style representation. We used these layer preferences and parameters $\mu=10^{2}$ for all the results.

In order to validate the effectiveness of the proposed method for controllable style transfer, we performed an experiment on \textbf{LPW dataset} \cite{LPW2016} which cover people with different ethicalities, a diverse set of everyday indoor and outdoor illumination environments, as well as natural gaze direction distributions.

In order to verify the effectiveness of the proposed method for gaze estimation, 3 public datasets were used to train the estimator with k-NN \cite{wang2017}, \textbf{MPIIGaze} dataset \cite{DBLP:journals/corr/abs-1711-09017} is used for test the accuacry. Three public datasets are:

\textbf{UTView} \cite{DBLP:conf/cvpr/SuganoMS14}: The data of subjects S0-S8 in UTView are used as subject 1--9 in our dataset. In total, there are 144 (head pose) $\times$ 160 (gaze directions) $\times$ 9 (subjects) $=$ 20,7360 training samples.

\textbf{SynthesEyes} \cite{DBLP:conf/iccv/WoodBZS0B15}: contains 11,382 synthesized close-up images of eyes. There are ten dynamic eye region model in this collection. The eye images are under a wide range of head poses, gaze directions, and illumination conditions.

\textbf{UnityEyes} \cite{DBLP:conf/etra/WoodBM0B16}: can rapidly synthesize large amounts of variable eye region images as training data. The model is based on high-resolution 3D face scans and uses real-time approximations for complex eyeball materials and structures as well as anatomically inspired procedural geometry methods for eyelid animation. Here, the dataset contains 28,332 synthetic eye images with different eye region model and eyeball materials.

\subsection{Qualitative Results}

To evaluate the qualification of our result, we compare proposed method with three state-of-the-art method, to compare the effective of proposed GAN with style transfer architecture, we compare with the \cite{DBLP:journals/corr/GatysEB15a} and Feifei Li et al.\cite{DBLP:conf/eccv/JohnsonAF16} which only use style transfer and \cite{DBLP:journals/corr/ShrivastavaPTSW16} which only use GAN. Basides that, we show the result of without modify generator and without modify discriminator. With all this five baseline method, we show the result of two different dataset which is UnityEyes \cite{DBLP:conf/etra/WoodBM0B16} and SynthesEyes \cite{DBLP:conf/iccv/WoodBZS0B15}. As Fig.8 and Fig.9 we can see that if closely observed, it can be seen that none of these styles has similar gaze angle with naturalistic images. The skin texture and the iris region in the refined synthetic images are qualitatively significantly more similar to the real images than to the synthetic images, it can be observed that the proposed method is more similar with real conditions by light and achieves outstanding results above \cite{DBLP:journals/corr/GatysEB15a} and Feifei Li et al.\cite{DBLP:conf/eccv/JohnsonAF16}. What's more, compare with without modify generator and without modify discriminator, the distribution of pupil and iris regions are dramatically clear.
\begin{figure}[!t]
\centering
\includegraphics[scale=.3]{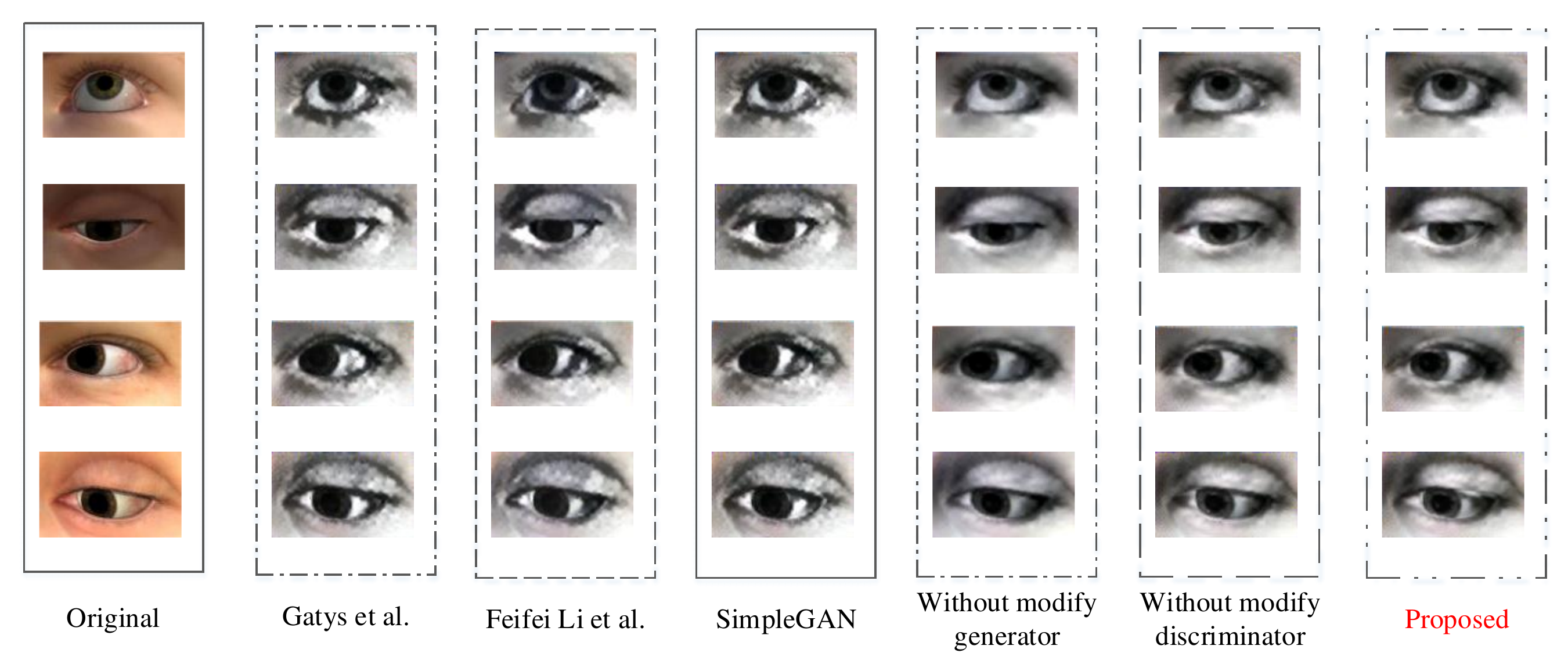}
\caption{Example output of proposed method for UnityEyes gaze estimation dataset.}
\end{figure}

\begin{figure}[!t]
\centering
\includegraphics[scale=.3]{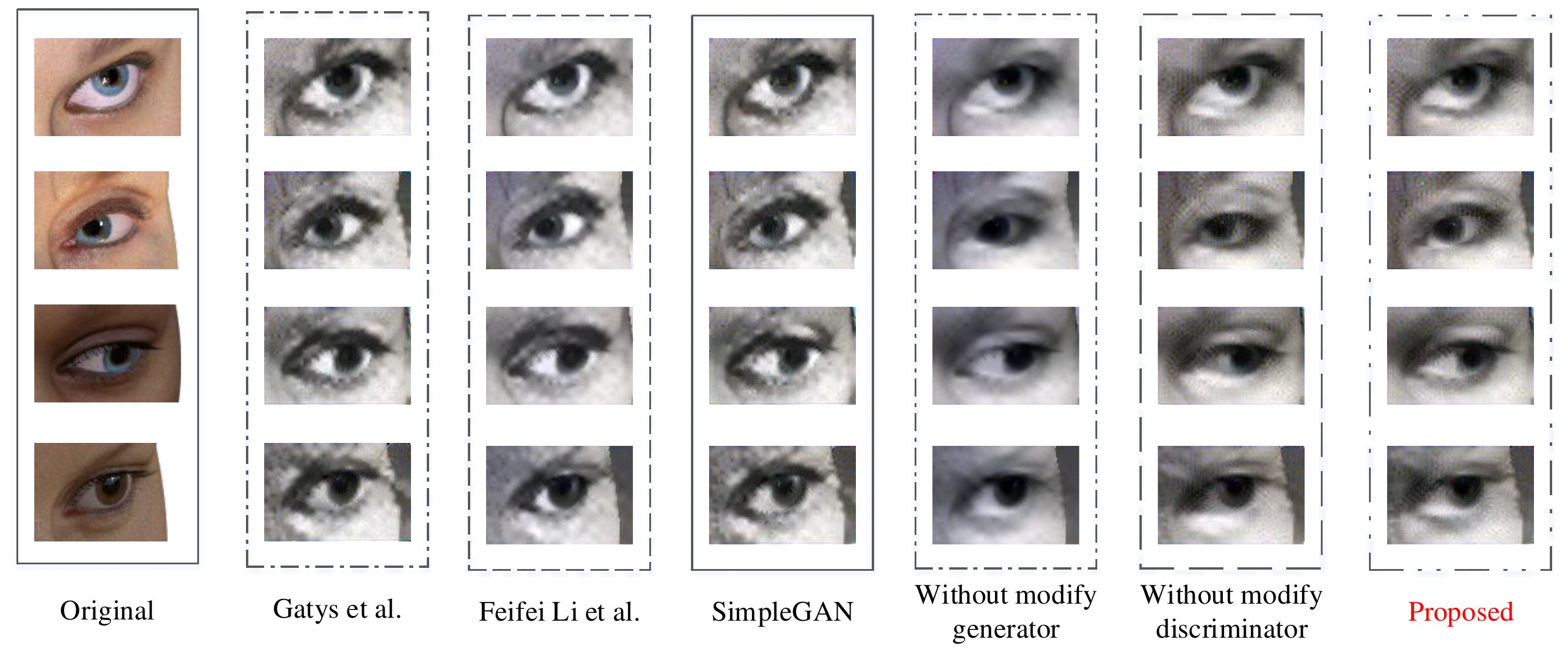}
\caption{Example output of proposed method for SynthesEyes gaze estimation dataset.}
\end{figure}

In order to validate the effectiveness of the proposed method, we compared it with available methods for several iteration in Fig.10 and Fig.11 on different dataset.  "Iter" means the number of iteration. Because \cite{DBLP:journals/corr/ShrivastavaPTSW16} is not stable so we only compare our method with \cite{DBLP:journals/corr/GatysEB15a} and Feifei Li et al.\cite{DBLP:conf/eccv/JohnsonAF16}, we can see that after iteration for serval iteration, proposed method can achieve stable distribution with less distortion, thus our result can be used as to train a stable gaze estimator.
\begin{figure}[!t]
\centering
\includegraphics[scale=.35]{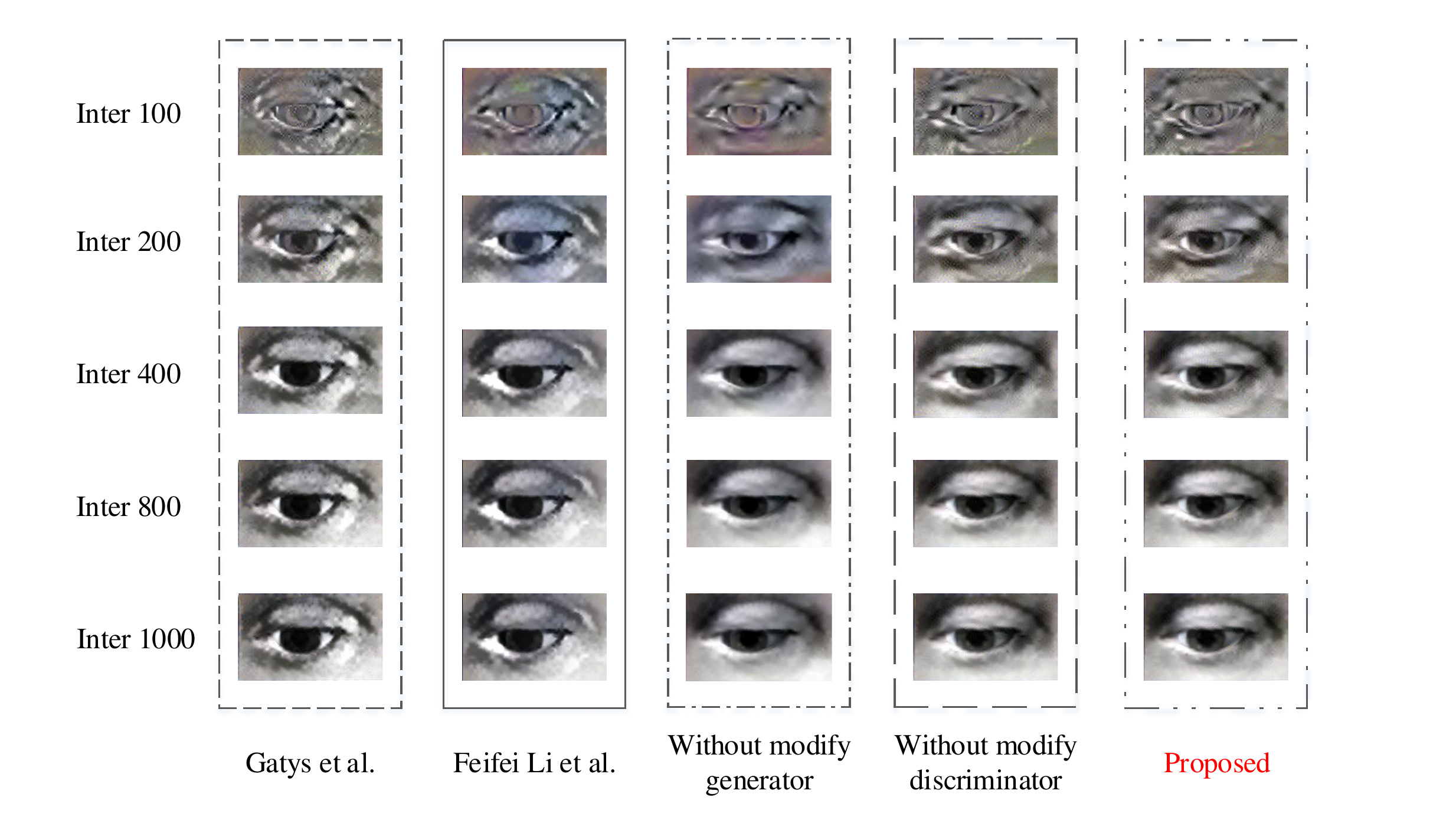}
\caption{Example output of proposed method for UnityEyes gaze estimation dataset for several iteration.}
\end{figure}

\begin{figure}[!t]
\centering
\includegraphics[scale=.35]{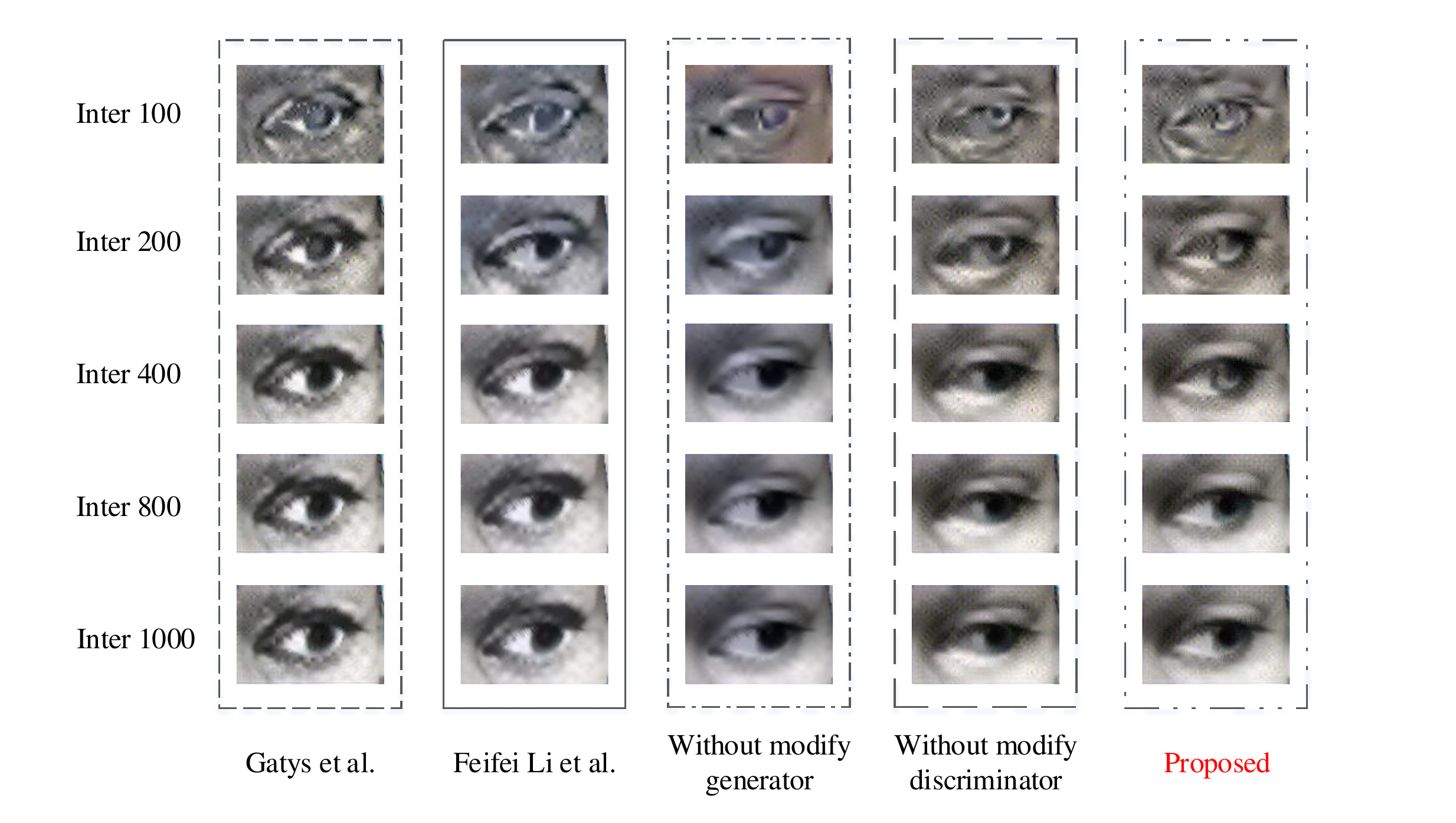}
\caption{Example output of proposed method for SynthesEyes gaze estimation dataset for several iteration.}
\end{figure}

\subsection{Appearance-based Gaze Estimation}
To verify the effectiveness of the proposed method, we perform experiments to assess both the quality of our refined images and their suitability for appearance-based gaze estimation.  We use COCO dataset to train the train net of coarse model net. And few of images from MPIIGaze dataset are chosen as target images. The gaze estimation dataset consists of 28,332 synthetic images from eye gaze synthesizer UnityEyes-fine dataset, six subjects of UTview datset and 350,428 real images from the MPIIGaze dataset. For UTview \cite{7299081}, the data of subjects S0, S2, S3, S4, S6 and S8 in UTView are used as subject 1--6 in our dataset. In total, there are 144 (head pose) $\times$ 160 (gaze directions) $\times$ 6 (subjects) $=$ 138,240 training samples and 8 (head pose) $\times$ 160 (gaze directions) $\times$ 6 (subjects) $=$ 7680  testing samples.
\begin{table}[!hbp]
\centering
\label{tabLr1}
\caption{Comparison of our method to the state-of-the-art on the part of MPIIGaze dataset of real eyes and UnityEyes-fine dataset. The third column indicates whether the methods are trained on Real/Synthetic data. The error means eye gaze estimation error in degrees.}
\begin{tabular}{|l|c|c|}
\hline
Method & Error & R/S \\
\hline
ALR \citep{DBLP:journals/pami/LuSOS14} & 16.7 & R\\
SVR \cite{6976920} & 16.6 & R\\
RF  \cite{6909631} & 15.4 & R\\
CNN with UT \cite{7299081} & 13.2 & R\\
K-NN with UT (ours) & 8.9 & R\\
CNN with UT (ours) & 10.2 & R\\
K-NN with Refined UnityEyes \\ \cite{DBLP:conf/iccv/WoodBZS0B15} & 10.2 & S\\
CNN with Refined UnityEyes \\ \cite{DBLP:conf/iccv/WoodBZS0B15} & 11.5 & S\\
CNN with Refined UnityEyes \\ (SimGANs \cite{DBLP:journals/corr/ShrivastavaPTSW16}) & 8.0 & S\\
K-NN with Refined UnityEyes(ours) & 8.3 & S\\
CNN with Refined UnityEyes(ours) & 7.7 & S\\
\hline
\end{tabular}

\end{table}

We evaluate the ability of our method for appearance-based gaze estimation from real dataset and synthetic image dataset. ALR \citep{DBLP:journals/pami/LuSOS14}, SVR \cite{6976920}, RF \cite{6909631}, convolution neural network \cite{Baltrusaitis}  and KNN \cite{DBLP:conf/iccv/WoodBZS0B15} are compared with our method as baseline methods. Similar to \cite{DBLP:conf/etra/WoodBM0B16}, we train a convolution neural network (CNN) to predict the eye gaze direction. For RF training, pixel-wise data is employed to represent the original eye image by converting it to column vector, the number of trees during training is set to $20$. For K-NN with UnityEyes refined images or UTview real images, considering that the computation cost increases with neighbor samples number, it can be found that a high-quality gaze estimator is obtained when the neighbor samples number is set to 50, which costs a shorter operating time. A comparison to the state-of-the-art can be shown in Table.1. Training the CNN on the refined images outperforms the state-of-the-art on the part of MPIIGaze dataset. We observe a large improvement in performance from training on the refined images and an significant improvement compared to the state-of-the-art.

\section{Conclusion}\label{sec:conclusion}

We propose a coarse-to-fine eye synthesis method through adversarial training to speed up refining synthetic images with less unlabeled real data. We make several key modifications to the GANs to make the net become an efficient refine model net to improve the suitability of gaze estimation and made the image not distorted. Comparing with the baseline methods, a large improvement in performance from training on the refined images is observed and the quantity of real data reduces by more than one order of magnitude.

\section*{ACKNOWLEDGMENTS}

The authors sincerely thank the editors and anonymous reviewers for the very helpful and kind comments to assist in improving the presentation of our paper. This work was supported in part by the National Natural Science Foundation of China Grant 61370142 and Grant 61272368, by the Fundamental Research Funds for the Central Universities Grant 3132016352, by the Fundamental Research of Ministry of Transport of P. R. China Grant 2015329225300.

\bibliographystyle{model2-names}
\bibliography{refs}

\end{document}